# UPDATING PROBABILITIES IN MULTIPLY-CONNECTED BELIEF NETWORKS

H.J. Suermondt and G.F. Cooper

Medical Computer Science Group
Stanford University
Stanford, California 94305-5479

*This paper focuses on probability updates in multiply-connected belief networks. Pearl has designed the method of conditioning, which enables us to apply his algorithm for belief updates in singly-connected networks to multiply-connected belief networks by selecting a loop-cutset for the network and instantiating these loop-cutset nodes. We discuss conditions that need to be satisfied by the selected nodes. We present a heuristic algorithm for finding a loop-cutset that satisfies these conditions.*

## 1. INTRODUCTION

Belief networks have proven to be a useful structure for representing knowledge in a probabilistic framework. A belief network is an acyclic, directed graph in which the nodes represent chance variables (e.g., in medical domains, a node could represent the presence vs. absence of a disease, or the possible values of a laboratory result). The arcs in a belief network represent the probabilistic influences between nodes. If a disease generally causes a symptom, then this could be represented by an arc from the disease to the symptom. An arc implies that it is possible to characterize the relationship between the connected nodes by a conditional probability matrix. See [1] for more details.

Pearl's algorithm for updating probabilities in singly-connected belief networks is well known [1]. We have implemented this algorithm recently. The main limitation of this algorithm is that its performance of belief updates in linear time is limited to singly-connected belief networks. A singly-connected belief network, also known as a causal polytree, has only a single pathway (in the undirected sense) from any node to any other node. A multiply-connected network, on the other hand, can have more than one pathway between nodes.

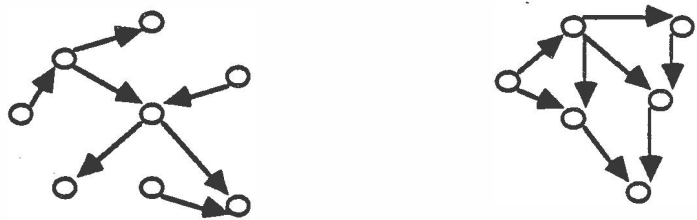

Figure 1.   Singly-connected belief network        Multiply-connected belief network

Unfortunately, many belief networks of practical use are multiply-connected. Pearl presents several ways to apply his algorithm to such networks [2],[3]. One, the method of conditioning, or reasoning by assumptions, provides a reasonable solution, provided that the network is not highly connected [3]. The method of conditioning is based on selection of a set of nodes, the loop-cutset, from the belief network and considering separately all possible combinations of values that these nodes can have. In other words, we treat each possible combination of values of the nodes of the loop-cutset as a separate case. Because we want to minimize the number of cases to consider, we are interested in finding the minimal loop-cutset: the set of nodes satisfying the requirements of the method of conditioning (described in Section 2) such that the product of the number of values of these nodes is minimal. The problem of finding the minimal loop-cutset is NP-hard (see Section 2.5); however, it is possible to find rapidly a small set of nodes such that, in many cases, it will be the minimal loop-cutset. In this paper, we shall describe a heuristic algorithm that will locate a set of nodes that can be used for the method of conditioning.

## 1.1 Notation and Terminology

When we refer to a node as a variable, we will use a capital letter ($X$); when we refer to a value of that node $X$, we will generally use lower-case $x$. This notation follows that used by Pearl in his detailed description of his algorithm [2].

For an arc from node $A$ to node $B$ we shall refer to node $A$ as the *parent* of node $B$, and to node $B$ as the *child* of node $A$. The parents of a node are all nodes from which there is a direct arc to that node; the set of children is defined similarly. The term *parent* is not to be confused with *ancestor*. Node $A$ is an *ancestor* of node $B$ if there is a directed path from node $A$ to node $B$. Similarly we can define the term *descendant* to mean the converse of *ancestor*: if node $A$ is an ancestor of node $B$, then node $B$ is a descendant of node $A$. We define the set of *neighbors* of a node as the union of the set of *parents* of that node and the set of *children* of the node.

Initially, the values of each node in the belief network are unspecified. When we receive information about a node that allows us to determine the node's value, that node becomes an *evidence node*. We use the terms *evidence nodes* and *observed nodes* interchangeably. After observing a node, we say that that node has been *instantiated*.

## 1.2. Pearl's Algorithm

When new evidence is observed for a proposition, we would like to propagate this evidence throughout the belief network in such a way that each proposition in the network is assigned a new measure of belief consistent with the axioms of probability theory. Pearl's algorithm performs this task through a series of local belief propagation operations, in which each node receives information messages from its neighboring nodes and combines these messages to update its measure of belief. Each node determines which of its neighbors need to receive updated information in order to maintain a correct probability distribution. Thus, through entirely local operations, the algorithm updates the probabilities for the nodes in the belief network to incorporate new evidence. More details can be found in [1] and [2].

An important advantage of Pearl's algorithm is that the effects of observations are propagated to all nodes in the belief network, rather than to a single node that is the object of a query. As a medical example, consider a belief network with several nodes that represent diseases and several nodes that represent symptoms, laboratory-test data, and other findings. Some belief-updating algorithms limit the transmission of information to a single query, such as one regarding the effect of observing a certain combination of symptoms on the probability distribution of a specific disease [4]. Only rarely, however, will the physician be interested in the probability of a single disease; rather, she will wish to consider a list of diseases that make up the differential diagnosis. Pearl's algorithm transmits the effects of observation of a given set of symptoms to all the disease nodes simultaneously, so queries regarding the effects of the same evidence on multiple propositions are efficient.

There are situations in which the structure of the network is such that observation of one node will have no effect on the probability distribution of some other nodes in the network. To prevent unnecessary calculations for nodes whose belief distributions will not be affected anyway, we can set the following blocking conditions for transmission of information:

- An evidence node does not send information from its children to its parents or from its parents to its children

- An evidence node does not send information from one child to any other children

- A node that has not been observed and that does not have any descendants that have been observed does not send information from one parent to any other parents

The last condition is an illustration of the property of belief networks of "independence except through links." If a node has not been observed, then all its parents will independently influence the probability distribution of that node, and information from one parent will not convey any information about the probabilities of the others; for an observed node, however, each parent node functions as a possible explanation for that observation. Therefore, information about the belief distribution of one parent will affect the distributions of the others.

For singly-connected networks, the structure of the belief network determines when to stop sending information. Even if no blocking conditions are observed, the worst that can happen is that the algorithm will do unnecessary work; the fact that a pathway is blocked means only that, beyond that block, no beliefs are changed by the new information, so the extra work of recalculating those beliefs is unnecessary. Because information is never sent back down an arc from which it arrived, belief propagation comes to a natural halt for singly-connected networks. As will be shown in Section 2.1, blocking conditions play a much more central role in belief updates for multiply-connected networks.

## 2. MULTIPLY-CONNECTED NETWORKS

As noted earlier, most belief networks created for practical purposes cannot be constrained to the singly-connected structure. Superimposing such a constraint could make the structure of the problem unnatural and counterintuitive. However, the most efficient form of Pearl's algorithm only applies to singly-connected networks; for multiply-connected networks, there are several related inference methods, all of which may suffer from combinatorial explosion since probabilistic inference using belief networks is known to be NP-hard [5]. The most generally applicable of these methods is the method of conditioning.

### 2.1 The Method of Conditioning

The method of conditioning is based on instantiating a small set of nodes to "cut" all loops in a multiply-connected belief network. A loop is a set of two undirected pathways between two nodes $X$ and $Y$ such that the pathways only intersect at $X$ and $Y$. By definition, loops only occur in multiply-connected networks. Because propagation of information is not centrally managed by Pearl's algorithm, when there are loops in a belief network, information may cycle infinitely. The method of conditioning prevents this cycling of messages by assuming that a select set of nodes, the loop-cutset, has been observed [3]. The nodes of the loop-cutset act as though they are evidence nodes, and thus prevent cycling of information by way of the blocking conditions described in Section 1.2.

A good way to look at belief updates using the method of conditioning is to act as though, rather than a single belief network, there is a collection of networks. The number of possible instantiations such that we cover every possible combination of value assignments to the members of the loop-cutset determines the number of copies of the belief network. These copies, which we have thus far called instantiations, all need to be processed when a new piece of information or evidence arrives.

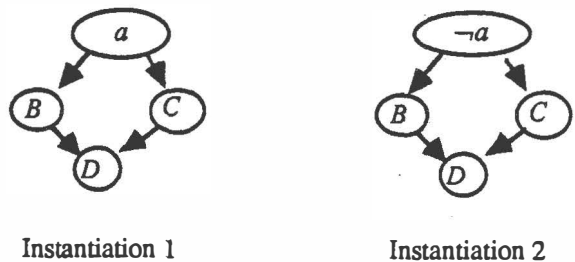

Instantiation 1　　　　Instantiation 2

**Figure 2.** How to consider multiple instantiations; Node $A$ has been assigned a different value in each instantiation, effectively stopping messages from $B$ through $A$ to $C$ and vice versa.

In Figure 2, if we assume that our loop-cutset consists of node $A$, then the copies of our network will be instantiation 1, in which we assume that node $A$ has value $a$, and instantiation 2, in which we assume that node $A$ is assigned $\neg a$. When we observe a new piece of evidence, the information in each network must be updated independently.

Thanks to the blocking conditions, the effects of new evidence can be calculated using Pearl's algorithm for singly-connected networks. The results of these calculations are then weighed by the joint probability of the nodes in the loop-cutset, given the observed evidence. The correctness of this approach is based on the rule of conditional probability: given evidence $E$ and a loop-cutset consisting of nodes $C_1...C_n$, then for any node $X$,

$$P(x \mid E) = \sum_{c_1...c_n} P(x \mid E, c_1...c_n) P(c_1...c_n \mid E) \qquad (1)$$

In the calculation for the new belief $P(x \mid E)$, the probability of $x$ given a certain instantiation of the loop-cutset nodes, $P(x \mid E, c_1...c_n)$, and the joint probability of that loop-cutset instantiation, $P(c_1...c_n \mid E)$, can both be calculated by Pearl's algorithm for singly-connected belief networks [3].

Intuitively, this method makes sense. When one is confronted with a problem that is too complex to handle, then an obvious strategy is to divide the problem into cases. We consider what would happen if we made certain simplifying assumptions; subsequently, we adjust the solution to take into account our original assumptions.

Unfortunately, this method may require a significant amount of computation time. For example, if our loop-cutset consists of just ten binary nodes, then we need $2^{10} = 1024$ terms in the above sum. The time complexity of this approach is exponential in the number of nodes in the loop-cutset. We can imagine highly connected belief networks, for which this method would not be practical due to the large size of the loop-cutset. However, provided that we can find a small loop-cutset for the multiply-connected network, the method of conditioning provides a workable and intuitively clear solution for belief networks with few loops.

## 2.2 Additional Problems in Multiply-Connected Networks

In addition to possible cycling of information, multiply-connected belief networks present another problem: parents of a node may share information from elsewhere in the network; therefore, they may not independently influence the probability distribution of their common child. Due to the local nature of belief propagations in Pearl's algorithm, this can lead to incorrect probability calculations unless the information shared by two parents is intercepted by a member of the loop-cutset. An example will clarify this problem.

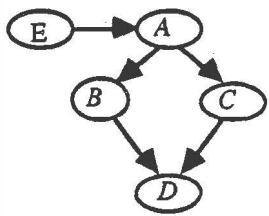

**Figure 3.** Example of a simple belief network

Consider the case given in Figure 3, where we are interested in the current probability of node $D$, given some information E that arrives at node $A$. We assume that nodes $A$ through $D$ are binary; for some node $X$, the possible values will be $x$ and $\neg x$. Assume nodes $A$ through $D$ have not been observed. Because node $D$ has not been observed, it will fulfill the blocking conditions; therefore, messages will not go around the loop formed by nodes $A$ through $D$, but rather, will stop propagation at node $D$. However, this example will show that unless node $A$, $B$ or $C$ is part of the loop-cutset, the results of top-down evidence propagation will not be correct according to the laws of probability, because evidence from node $E$ can reach node $D$ through multiple pathways.

If none of nodes $A$, $B$ and $C$ are in the loop-cutset, when some new evidence E is observed, we first use it to calculate the new belief distribution for node $A$. Next, the evidence is propagated to nodes $B$ and $C$.

338

Node $B$ and node $C$ send the information to node $D$. Node $D$ now reads the messages from $B$ and $C$ and calculates its current belief.

Because of the local nature of belief propagation in Pearl's algorithm, the probabilities for node $D$ are calculated as follows:

$$P(d \mid E) = \begin{aligned} & P(d \mid b, c) P(b \mid E) P(c \mid E) \\ & + P(d \mid b, \neg c) P(b \mid E) P(\neg c \mid E) \\ & + P(d \mid \neg b, c) P(\neg b \mid E) P(c \mid E) \\ & + P(d \mid \neg b, \neg c) P(\neg b \mid E) P(\neg c \mid E) \end{aligned} \quad (2)$$

$P(b \mid E)$ and $P(c \mid E)$ are calculated locally, however, as follows:

$$P(b \mid E) = P(b \mid a) P(a \mid E) + P(b \mid \neg a) P(\neg a \mid E)$$

$$P(c \mid E) = P(c \mid a) P(a \mid E) + P(c \mid \neg a) P(\neg a \mid E)$$

Analogously, we can calculate $P(\neg b \mid E)$ and $P(\neg c \mid E)$. After substituting these into Equation (2), and collecting terms, we get an equation that contains several terms that would not be part of the equation if we were to follow the axioms of probability. An example of such an incorrect term is

$$P(d \mid b, c) P(b \mid a) P(a \mid E) P(c \mid \neg a) P(\neg a \mid E)$$

Note that this term contains the probabilities $P(a \mid E)$ and $P(\neg a \mid E)$ which are logically inconsistent with one another. Due to the local nature of belief propagation, however, these terms are unavoidable unless we condition on node $A$, $B$ or $C$. If we fail to do this, we get cross-multiplications of incompatible terms.

On the other hand, if we make node $A$ part of our loop-cutset, then we will consider separately the case where node $A$ has value $a$ and the case where node $A$ has value $\neg a$. The belief distribution for node $D$ would be calculated as a combination of these cases:

$$P(d \mid E) = P(d \mid a, E) P(a \mid E) + P(d \mid \neg a, E) P(\neg a, E) \quad (3)$$

In this equation, $P(d \mid a, E)$ and $P(d \mid \neg a, E)$ are the calculated beliefs for each instantiation of the loop-cutset; $P(a \mid E)$ and $P(\neg a \mid E)$ are the cutset weights for these instantiations. To simulate this calculation using Pearl's algorithm, we calculate $P(d \mid a, E)$ and $P(d \mid \neg a, E)$ as follows:

$$P(d \mid a, E) = \begin{aligned} & P(d \mid b, c) P(b \mid a, E) P(c \mid a, E) \\ & + P(d \mid b, \neg c) P(b \mid a, E) P(\neg c \mid a, E) \\ & + P(d \mid \neg b, c) P(\neg b \mid a, E) P(c \mid a, E) \\ & + P(d \mid \neg b, \neg c) P(\neg b \mid a, E) P(\neg c \mid a, E) \end{aligned} \quad (4)$$

We can calculate $P(d \mid \neg a, E)$ in an analogous manner.

We assume that node $A$, as a member of the loop-cutset, is an evidence node; therefore, due to the blocking conditions, $P(b \mid a, E) = P(b \mid a)$. Analogously, $P(c \mid a, E) = P(c \mid a)$, etc. After substituting these simplifications into Equation (4) (and its analogs), and substituting Equation (4) (and its analogs) into Equation (3), we have the following result:

$$P(d \mid E) = \begin{aligned} & P(d \mid b, c) P(b \mid a) P(c \mid a) P(a \mid E) \\ & + P(d \mid b, c) P(b \mid \neg a) P(c \mid \neg a) P(\neg a \mid E) \\ & + P(d \mid b, \neg c) P(b \mid a) P(\neg c \mid a) P(a \mid E) \\ & + P(d \mid b, \neg c) P(b \mid \neg a) P(\neg c \mid \neg a) P(\neg a \mid E) \\ & + P(d \mid \neg b, c) P(\neg b \mid a) P(c \mid a) P(a \mid E) \\ & + P(d \mid \neg b, c) P(\neg b \mid \neg a) P(c \mid \neg a) P(\neg a \mid E) \\ & + P(d \mid \neg b, \neg c) P(\neg b \mid a) P(\neg c \mid a) P(a \mid E) \\ & + P(d \mid \neg b, \neg c) P(\neg b \mid \neg a) P(\neg c \mid \neg a) P(\neg a \mid E) \end{aligned} \quad (5)$$

Equation (5) is the result of calculating the belief for $d$ given evidence E according to Pearl's algorithm provided we add node $A$ to our loop-cutset. Equation (5) is consistent with the axioms of probability theory. We reach the same result if we use node $B$ or $C$ as our cutset node, instead of node $A$.

### 2.3 A Condition for the Loop-Cutset Nodes

The previous example shows that the nodes of the loop-cutset must not only stop infinite cycling of information, but also enable the local probability calculations to achieve correct results. We thus conclude that when we adjust Pearl's algorithm for multiply-connected belief networks, in order to get belief calculations consistent with the axioms of probability, the loop-cutset must satisfy the following **Loop-Cutset Condition:**

> The loop-cutset must contain at least one node from every loop in the belief network such that this node is a child to no more than one other node in the same loop

If we do not add at least one node from every loop in the network to our loop-cutset, we may fail to prevent cycling of information. If the only loop-cutset node in a certain loop is a child to more than one other node in that loop, then it receives top-down information more than once, leading to the incorrect belief updates demonstrated in Section 2.2. The algorithm described in Section 2.4 finds a loop-cutset that satisfies these conditions.

### 2.4 An Algorithm for Finding the Appropriate Loop-Cutset

It is important to have as small a loop-cutset as possible in order to minimize the number of possible instantiations of the loop-cutset nodes. The following algorithm creates a loop-cutset that satisfies the loop-cutset condition. It is heuristic in the sense that it attempts to find a small loop-cutset, but it does not guarantee that the minimal set will always be found. Its main steps are:

1. Remove all parts of the network that are not in any loop.

2. If there are any nodes left, find a good loop-cutset candidate. A good loop-cutset candidate is defined as a node that satisfies the loop-cutset condition and the heuristic criteria described below. Add this node to the loop-cutset, then remove it from the network; return to 1. Terminate when no nodes remain in the network.

Step **1** is based on the fact that we only want to add nodes to our loop-cutset if those nodes are part of at least one loop. No singly-connected nodes of the belief network will be members of the loop-cutset; therefore, all singly-connected parts can be deleted. We start this process by finding any nodes that have a single parent and no children, or a single child and no parents; in other words, we find nodes that have only a single neighbor. These nodes are not part of any loop, so we can remove them from our network, because they do not need to be part of the loop-cutset. We also remove the arcs that connect each pruned node to its single neighbor. After removing each node and its arc, we consider its neighbor. If this neighbor now also meet the condition for removal (i.e., it has a single neighbor), then we can repeat the process. We continue until no nodes with a single neighbor remain in the network.

Let us illustrate this for the network in Figure 4 (A). Node $A$ has only one neighbor; it is therefore not part of any loop, so we can remove it. We follow its arc to node $C$, which now has two neighbors, since node $A$ has been deleted. If a node has more than one neighbor, we do not know whether it is part of a loop; for example, both node $C$ and node $E$ have two neighbors; node $C$ is not part of any loop, but node $E$ is. Therefore, we leave node $C$; first, we see whether there is another node we can remove. In this case, node $B$ also has only a single neighbor. Therefore, we delete node $B$ and follow its arc, returning to node $C$. This time, node $C$ has only one neighbor left, because node $A$ and node $B$ have been deleted, so now we know node $C$ is not part of a loop and we can delete it. We follow its arc to node $D$. Node $D$ has two neighbors, so we cannot continue. There are no other nodes with only a single neighbor; therefore, we have completed step 1. At this point, all singly-connected parts of the belief network have been removed; all nodes remaining in the network are members of one or more loops.

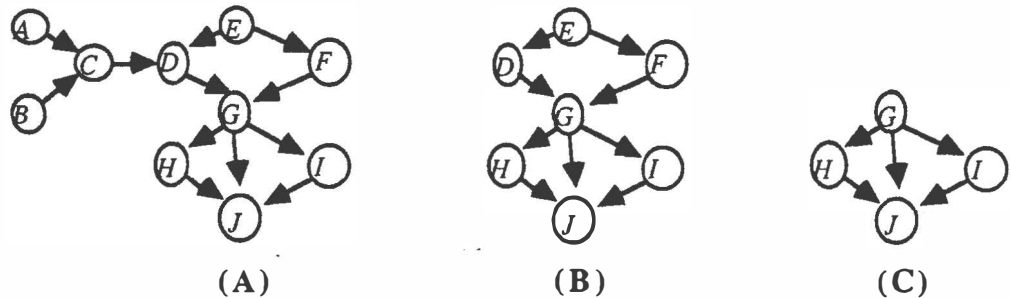

Figure 4. Belief network to illustrate the heuristic algorithm; A: the original network; B: the original network after removal of all singly-connected parts; C: what is left of the network after a loop has been removd

The goal of step 2 is to find a node that satisfies the loop-cutset condition (see Section 2.3) for as many loops as possible in order to minimize the number of possible instantiations of the loop-cutset nodes. Therefore, in step 2 we employ a heuristic strategy that has the following three components. First, we only consider nodes with one or fewer parents. Thus, we avoid nodes that violate the loop-cutset condition by having more than one parent in the same loop. Second, of the nodes that remain under consideration we select the node that has the most neighbors. Since all nodes remaining in our network are members of one or more loops, the number of loops that a node is part of will increase with the number of neighbors that that node has. By adding the node with the most neighbors, we hope to minimize the number of nodes that need to be added to the loop-cutset to satisfy the loop-cutset condition. Third, if there is a choice between multiple nodes that each have the same maximum number of neighbors, then we select the node with the fewest values. This will minimize the number of possible value assignments to the loop-cutset nodes. If there are large discrepancies between the number of values of cutset candidates, then one might consider giving this criterion priority over the neighbor maximization criterion; one might also combine the two criteria into a single weight function. Both criteria attempt to minimize the number of instantiations of the loop-cutset nodes, but if the nodes are similar in number of values, the neighbor maximization criterion is likely to be preferable.

After heuristically deciding which node would be a good loop-cutset member, we add this node to our set and remove it from the network. By removing a node that is in a loop, we potentially make the remainder of that loop singly-connected. Because we may thus create new singly-connected parts of the network, we need to return to step 1 and remove these parts.

In Figure 4 (B), nodes $G$ and $J$ have more than one parent so they cannot be considered for the loop-cutset. The nodes remaining under consideration all have two neighbors and are therefore equally attractive candidates by the first two criteria. If, for example, node $E$ has fewer values than any other node, by the fewest-values criterion we decide to add it to to our loop-cutset. After removing node $E$, we can prune its neighbors, node $D$ and node $F$, since these nodes both have only one remaining neighbor. After removing these nodes, once more all remaining nodes have more than one neighbor, so all nodes are members of one or more loops. Therefore, we need to look for the best loop-cutset candidate again.

With what is left of the network (Figure 4 (C) ), we repeat the same process. It is now clear that node $G$ has the most neighbors of all nodes with one or fewer parents, so we select it next. After adding node $G$ to the set and deleting it from our network, we follow its arcs to nodes $H$ and $I$. These nodes now have only one neighbor left, so we can remove them. For either one of them, we follow the arc to node $J$, which has no remaining neighbors. We remove node $J$, and are finished. Our final loop-cutset consists of nodes $E$ and $G$.

341

## 2.5 Computational Complexity

It is known that general probabilistic inference using belief networks is NP-hard [5]; since the method of conditioning is a generally applicable inference algorithm for belief networks, it is to be expected that inference using this method is of exponential time complexity with respect to the number of nodes in the network. In particular, the average time complexity of a single probability update using the method of conditioning is proportional to the product of the number of values of the nodes in the loop-cutset.

The exponential nature of the method of conditioning makes it important that we find a small loop-cutset; if possible, we would like this loop-cutset to be minimal. As mentioned earlier, however, finding the minimal loop-cutset is also an NP-hard problem. We can show that the minimal loop-cutset (MLC) problem is NP-hard by reducing the minimal vertex cover (MVC) problem to MLC. The MVC problem, which is known to be NP-hard [7], is defined as follows: Given a finite, undirected graph (V, E) with vertices V and edges E, find a subset V' of V that is of smallest cardinality such that for all edges e in E, E has an endpoint in V'. Informally, the reduction from MVC to MLC is as follows. Each edge $(V_i, V_j)$ in MVC induces the following structure in MLC:

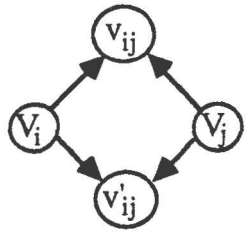

Figure 5. Portion of a belief network in the MLC problem corresponding to a link between nodes $V_i$ and $V_j$ in the MVC problem.

In Figure 5, $V_i$, $V_j$, $v_{ij}$, $v'_{ij}$ are propositional variables, and $v_{ij}$ and $v'_{ij}$ are instantiated to either T or F. The reduction thus constructs a loop in MLC for each edge in MVC. By the loop-cutset condition outlined in Section 2.3, for each such loop in MLC we must include either $\{V_i\}$, $\{V_j\}$, or $\{V_i, V_j\}$ in our loop-cutset, which corresponds directly to finding a cover for edge $(V_i, V_j)$ in MVC. For example, a simple four node problem instance of MVC is reduced to a problem instance of MLC as follows: (see Figure 6)

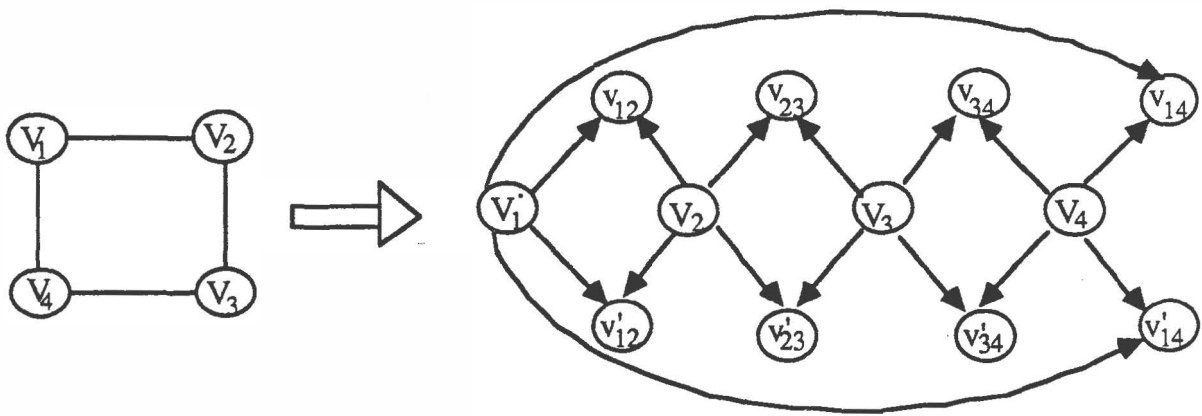

Figure 6. Illustration of the reduction of a simple instance of the MVC problem in an undirected graph to an instance of the MLC problem in a belief network.

In figure 6, for example, $\{V_1, V_3\}$ is a minimal loop-cutset in MLC and a minimal node cover in MVC, but $\{V_1, V_2\}$ is not. It is clear that a minimal loop-cutset for all loops in such an MLC construction is a minimal covering for all vertices in the corresponding MVC problem. Thus, the problem of finding the minimal loop-cutset (MLC) is NP-hard.

Because of the computational complexity of the MLC problem, we have developed the heuristic algorithm described in Section 2.4 to find a loop-cutset that is generally small, but that is not guaranteed to be minimal. The worst case time complexity for finding a loop-cutset using this algorithm is $O(n^2)$. We have implemented this algorithm for a general belief network tool called KNET [6]. We are currently in the process of formally evaluating the efficiency of the heuristic. Our preliminary results indicate that our algorithm performs well at finding a small loop-cutset which, in turn, allows us to apply Pearl's method of conditioning to probabilistic inference for a select class of belief networks.

## Acknowledgements

We wish to thank Lyn Dupre, Eric Horvitz, Ross Schachter, Ramesh Patil, and an anonymous referee for useful comments on earlier versions of this document. Support for this work was provided by the National Science Foundation under grant IRI-8703710, and the National Library of Medicine under grant R01-LM04529. Computer facilities were provided by the SUMEX-AIM resource under grant RR-00785 from the National Institutes of Health.

## References

[1] Pearl, J., Fusion, propagation and structuring in belief networks. *Artificial Intelligence*, 29 (1986), pp.241-288.

[2] Pearl, J., Distributed revision of composite beliefs. *Artificial Intelligence*, 33 (1987), pp. 173-215.

[3] Pearl, J., A constraint-propagation approach to probabilistic reasoning, in: Kanal, L.N. and Lemmers, J.F., eds., 1986, Uncertainty in Artificial Intelligence, Elsevier Science Publishers, Amsterdam, The Netherlands, pp. 357-369.

[4] Shachter, R. D., Evaluating influence diagrams. *Operations Research*, 34 (1986), pp. 871-882.

[5] Cooper, G. F., The computational complexity of probabilistic inference using belief networks. Memo KSL-87-27, Knowledge Systems Laboratory, Stanford University, Stanford, CA, May 1988.

[6] Chavez, R.M. and Cooper, G.F., KNET: Integrating hypermedia and normative Bayesian modeling. *Proceedings of the AAAI Workshop on Uncertainty in Artificial Intelligence*, 1988.

[7] Garey. M.R. and Johnson, D.S., Computers and Intractability: A Guide to the Theory of NP-Completeness, W.H.Freeman and Company, San Francisco, 1979, p. 190.